\documentclass[runningheads]{llncs}
\usepackage[T1]{fontenc}
\usepackage{graphicx}
\usepackage{amsmath}
\usepackage{siunitx}
\usepackage{listings}
\usepackage{hyperref}

\usepackage{amssymb}
\usepackage{caption}
\usepackage[numbers]{natbib}  
\usepackage{booktabs}
\usepackage{comment}
\usepackage{enumitem}
\usepackage{subcaption}

\usepackage{tikz}
\usepackage{textcomp}
\usetikzlibrary{arrows.meta, positioning, decorations.pathreplacing}
\usepackage{bbm}

\renewcommand{\mid}[0]{\hspace{0.05 cm}|\hspace{0.05 cm}}

\begin{document}
\title{V-CEM: Bridging Performance and Intervenability in Concept-based Models}
%
%
\author{Francesco De Santis\inst{1}
\and
Gabriele Ciravegna\inst{1,2}
\and
Philippe Bich\inst{1}
\and 
Danilo Giordano\inst{1}
\and 
Tania Cerquitelli\inst{1}
\thanks{Paper accepted at \textit{The $3^{rd}$ World Conference on Explainable Artificial Intelligence}.}
}
\authorrunning{F. De Santis et al.}
%
\institute{Politecnico di Torino, Turin, 10129, Italy \\
\email{\{name.surname\}@polito.it} \and
Centai Institute, Turin, 10138, Italy}
 \maketitle              
\begin{abstract}

Concept-based eXplainable AI (C-XAI) is a rapidly growing research field that enhances AI model interpretability by leveraging intermediate, human-understandable concepts. This approach not only enhances model transparency but also enables human intervention, allowing users to interact with these concepts to refine and improve the model's performance. Concept Bottleneck Models (CBMs) explicitly predict concepts before making final decisions, enabling interventions to correct misclassified concepts. While CBMs remain effective in Out-Of-Distribution (OOD) settings with intervention, they struggle to match the performance of black-box models. Concept Embedding Models (CEMs) address this by learning concept embeddings from both concept predictions and input data, enhancing In-Distribution (ID) accuracy but reducing the effectiveness of interventions, especially in OOD scenarios. In this work, we propose the Variational Concept Embedding Model (V-CEM), which leverages variational inference to improve intervention responsiveness in CEMs. We evaluated our model on various textual and visual datasets in terms of ID performance, intervention responsiveness in both ID and OOD settings, and Concept Representation Cohesiveness (CRC), a metric we propose to assess the quality of the concept embedding representations. The results demonstrate that V-CEM retains CEM-level ID performance while achieving intervention effectiveness similar to CBM in OOD settings, effectively reducing the gap between interpretability (intervention) and generalization (performance). 

\end{abstract}

\section{Introduction}
Concept-Bottleneck Models (CBMs)~\cite{koh2020concept} have emerged as a promising approach to interpretable machine learning by making task predictions through intermediate, human-understandable concepts. This architecture enhances model transparency by providing insight into the decision-making process through an interpretable mapping between concepts and outputs. Additionally, CBMs offer a distinctive advantage: the ability for human users to \textit{intervene} on the intermediate concept predictions. This allows users both to rectify misclassified concepts, improving model performance, and to gain a deeper understanding of the relationships between concepts and task labels.

However, CBMs struggle with generalization, exhibiting limited performance.
Their performance is constrained by the intermediate bottleneck, which restricts their ability to match the predictive accuracy of black-box models that directly map inputs to outputs. To address this issue, Concept Embedding Models~\citep{zarlenga2022concept, kim2023probabilistic} (CEMs) have been introduced. CEMs generate dedicated embedding representations for each concept, thus alleviating the constrained representational capacity of the concept bottleneck. This approach improves model performance achieving black-box accuracy, while preserving a degree of intervenability (i.e., the level of efficacy of intervention) and interpretability. 
Besides testing model intervenability in In-Distribution (ID) settings, in this paper we propose testing model intervenability in Out-of Distribution scenarios (OOD).    
Our experiments show that the CBM architecture remains responsive to interventions on concept representations in both ID and OOD settings. In contrast, CEM exhibits very limited intervenability in OOD scenarios. Theoretically, this is due to CBM relying exclusively on predicted concepts for final decisions, whereas CEMs predictions are based on concept embeddings, which integrate both concept predictions and raw input data. This entanglement negatively impacts CEM’s intervenability in OOD settings.
To address this challenge, we propose the Variational Concept Embedding Model (V-CEM), which utilizes variational inference to achieve black-box-level accuracy on ID tasks, while maintaining high intervention responsiveness in both ID and OOD scenarios.  

In summary, this work makes the following key contributions: i) We demonstrate that while CEMs can achieve higher ID accuracy compared to CBMs, their ability to support interventions in OOD scenarios is significantly limited; ii) We introduce V-CEM, a model that achieves black-box generalization performance under ID conditions, comparable to CEMs; iii) We show that V-CEM retains responsiveness to interventions in both ID and OOD scenarios, similar to CBMs.

The manuscript is structured as follows. In Section~\ref{sec:background}, we provide the foundational concepts necessary to understand this work. Section~\ref{sec:var_cem} introduces V-CEM, while Section~\ref{sec:eval} outlines the metrics used to evaluate concept representations. In Section~\ref{sec:exp}, we present the results of our experimental campaign. Finally, in Section~\ref{sec:related}, we review related works, and Section~\ref{sec:conc} offers concluding remarks. Code is publicly available\footnote{ \href{https://github.com/francescoTheSantis/Variational-Concept-Embedding-Model}{https://github.com/VCEM}}.

\section{Background}
\label{sec:background}
\subsubsection{Concept Bottleneck Models (CBMs).} Let $x \in X \subset \mathbb{R}^d$ be an input realization, $c \in C \subset [0,1]^k$ represent interpretable concepts, and $y \in Y \subset \{0, \dots, N\}$ denote the task label. CBMs assume a generative process where $x$ determines $c$, which in turn influences $y$. A CBM consists of a concept encoder $p(c \mid x)$ and a task classifier $p(y \mid c)$, trained end-to-end to approximate $p(y,c\mid x)=p(y\mid c)p(c\mid x)$. The corresponding Probabilistic Graphical Model (PGM) is shown in Figure~\ref{fig:pgm}a. Modifying a concept $c_j$ removes its reliance on $x$. This characteristic is especially crucial in OOD scenarios, as it enables to completely replace the concept representation generated by the concept encoder for a given concept. However, the bottleneck on $c$, while enhancing interpretability, limits performance in ID settings, resulting in a trade-off between interpretability and accuracy.

\subsubsection{Concept Embedding Models (CEMs).} 
\begin{figure}[t]
    \centering
    \begin{subfigure}[b]{0.3\textwidth}
        \centering
        \begin{tikzpicture}[
            node distance=0.7cm, 
            roundnode/.style={circle, draw=black, minimum size=0.5cm, font=\small},
            ]
            
            \node[roundnode] (x) {\(X\)};
            \node[roundnode, right=of x] (c) {\(C\)};
            \node[roundnode, right=of c] (y) {\(Y\)};
            
            \draw[->] (x) -- (c);
            \draw[->] (c) -- (y);
    
        \end{tikzpicture}
        \caption{CBM}
        \label{fig:cbm_pgm}
    \end{subfigure}
    \hfill
    \begin{subfigure}[b]{0.3\textwidth}
        \centering
        \begin{tikzpicture}[
            node distance=0.7cm, 
            roundnode/.style={circle, draw=black, minimum size=0.5cm, font=\small},
            ]
            
            \node[roundnode] (x) {\(X\)};
            \node[roundnode, right=of x] (c) {\(C\)};
            \node[roundnode, below=of c] (c_) {\(\mathbf{C}\)};        
            \node[roundnode, right=of c_] (y) {\(Y\)};
            
            \draw[->] (x) -- (c);
            \draw[->] (c) -- (c_);
            \draw[->] (c_) -- (y);
            \draw[->] (x) -- (c_);
    
        \end{tikzpicture}
        \caption{CEM}
        \label{fig:cem_pgm}
    \end{subfigure}
    \hfill
    \begin{subfigure}[b]{0.3\textwidth}
        \centering
        \begin{tikzpicture}[
            node distance=0.7cm, 
            roundnode/.style={circle, draw=black, minimum size=0.5cm, font=\small},
            ]
            
            \node[roundnode] (x) {\(X\)};
            \node[roundnode, right=of x] (c) {\(C\)};
            \node[roundnode, below=of c] (c_) {\(\mathbf{C}\)};        
            \node[roundnode, right=of c_] (y) {\(Y\)};
            
            \draw[->] (x) -- (c);
            \draw[->] (c) -- (c_);
            \draw[->] (c_) -- (y);
            \draw[dotted, ->] (x) -- (c_);
            \draw[dotted, ->] (c) to[out=320, in=45] (c_);

        \end{tikzpicture}
        \caption{V-CEM}
        \label{fig:v_cem_pgm}
    \end{subfigure}
    \caption{Probabilistic Graphical Models of a) CBMs, b) the CEMs, and c) the proposed V-CEM architecture. Solid lines represent the data generation process, while dotted lines represent inference.}
    \label{fig:pgm}
\end{figure}
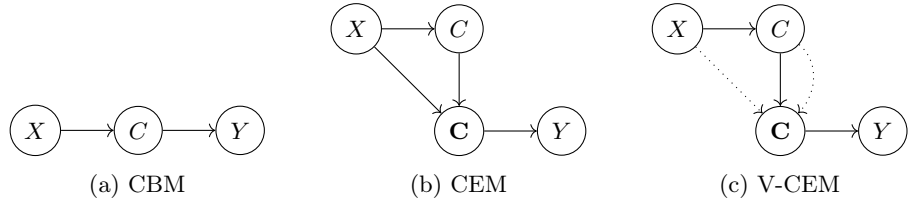

CEM alleviates the usual conflict between interpretability and performance by introducing a rich concept representation, the concept embedding $\mathbf{c} \in \mathbf{C} \subset \mathbb{R}^{k \times m}$, as shown in CEM PGM in Figure~\ref{fig:pgm}b. 
Unlike the CBM architecture, CEM defines a new conditional distribution $p(\mathbf{c} \mid c, x)$ that integrates both the input $x$ and the concept $c$, enabling the generation of concept embeddings that capture concept-specific information enriched by the input instance $x$. These embeddings are then used to model the distribution $p(y \mid \mathbf{c})$, which predicts task labels. Similarly to CBMs, CEM is trained to approximate $p(y,c\mid x)$.  Despite utilizing embeddings, CEM maintains the ability to support concept interventions: modifying a concept influences the conditional distribution $p(\mathbf{c} \mid x, c)$, thereby altering the generated embeddings. The dependence on $x$, which contributes to high ID performance, still remains after human intervention. The reliance on $x$, which contributes to strong ID performance, persists even after human intervention. As a result, CEM becomes less responsive to interventions in OOD scenarios, as the concept embedding generated in these cases may contain poor-quality information that cannot be overridden by human input.

\subsubsection{Intervention. }
\label{sec:emb_int}

Interventions in Concept based Models enable humans to correct model errors and gain insights into the relationship between concepts and tasks. This capability is crucial for developing interpretable models by improving transparency, trust, and control over decision-making. For instance, in a classification task where the objective is to categorize birds based on a set of concepts representing their features, if the concept \textit{Red Breast} of an image depicting a \textit{Red Breasted Parrot} is misclassified, the model might assign an incorrect bird label to the image. A human can adjust the concept prediction, which in turn may alter the final task prediction of the model. 
Different approaches enable various types of interventions: concept intervention~\cite{koh2020concept, zarlenga2022concept}, where the predicted concept is directly replaced, and concept embedding intervention~\cite{kim2023probabilistic}, where the concept's embedding is adjusted. Formally, in concept intervention, the concept $c_j \sim p(c_j \mid x)$ is replaced with $c_j := c'_j$, where $c'_j$ is the concept assigned by the human. In a similar manner, in concept embedding intervention, $\mathbf{c}_j \sim p(\mathbf{c}_j \mid x)$ is substituted with $\mathbf{c}_j := \mathbf{c}'_j$, where $\mathbf{c}'_j$ is the embedding representing concept $j$ that the human uses to correct the misclassified concept.

\section{Variational CEM} 
\label{sec:var_cem}
We propose Variational CEM, a methodology to maintain CEM performance in ID settings by leveraging the rich, sample-specific information of the concept embeddings while ensuring their dependence primarily on the underlying concepts. At the same time, V-CEM enables targeted interventions on the concept embeddings that completely override their dependency on the input, ensuring high intervenability also in OOD scenarios. In Section~\ref{sec:v-cem_arch} we describe V-CEM architecture, while in Section~\ref{sec:v-cem_train} we describe its training.

\subsection{V-CEM Architecture}
\label{sec:v-cem_arch}
\begin{figure}[t]
\centering
\includegraphics[width=\textwidth]{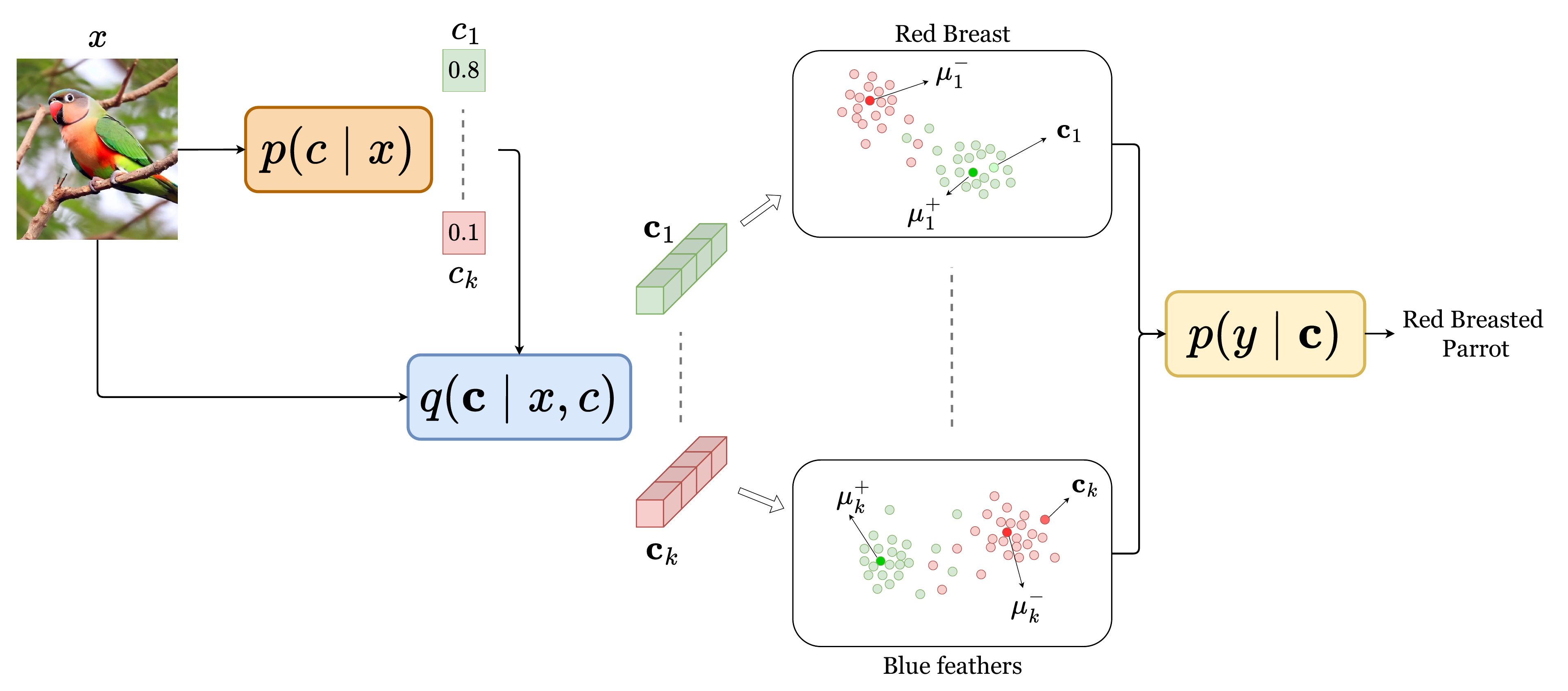}
\caption{Illustration of the V-CEM architecture. Given an image of a parrot with a red breast, V-CEM concept encoder $p(c|x)$ assigns a high probability to the ``Red Breast'' concept and a low probability to ``Blue Feathers'', which is absent. The approximate posterior $q(\mathbf{c}|x,c)$ maps concept prediction to concept embeddings clustered around $\mu^+_{\text{Red Breast}}$ and $\mu^-_{\text{Blue Feathers}}$, respectively. These embeddings are then employed to condition $p(y \mid \mathbf{c})$ and enable a correct label prediction (``Red Breasted Parrot'').}
\label{fig:schema}
\end{figure}

 As shown in Figure~\ref{fig:schema}, V-CEM is composed first of a concept encoder $p(c|x)$, mapping the input data $x$ to an intermediate, interpretable concept layer $c$. Concept embeddings, $\mathbf{c}$, are generated from $q(\mathbf{c}\mid x, c)$ using both concept predictions and input features. The classification head $p(y|\mathbf{c})$ works on the concept embeddings to produce the final class prediction $y$.
 
However, from a probabilistic point of view, we assume a generative process where the concept embeddings $\mathbf{c}$ are only influenced by the interpretable concepts $c$ and not by the input $x$, which is only used to derive the concept $c$. Similarly to CEM, the task label $y$ is generated from a distribution conditioned on the concept embeddings. The PGM corresponding to this formulation is depicted by the solid lines in Figure~\ref{fig:pgm}c. This generative framework leads to the following factorization:  

\begin{equation}
    p(x,c,\mathbf{c},y) = p(x)p(c|x)p(\mathbf{c}|c)p(y|\mathbf{c})
    \label{eq:factorization}
\end{equation}

With respect to the CEM architecture, we introduce a prior $p(\mathbf{c} \mid c)$, which we will discuss in detail later. Also, notice how the concept embedding probability is only conditioned by the concept predictions $p(\mathbf{c}|c)$. 
Similarly to CBM and CEM, our objective is to approximate the joint distribution $p(y, c \mid x)$. Since $\mathbf{C}$ is unobservable, we account for its effect on the relationships between $X$, $C$, and $Y$ by marginalizing over all possible values of $\mathbf{C}$:
{\small
\begin{align}
    p(c,y|x) =& \int_{\mathbf{C}} \frac{p(x,c,\mathbf{c},y)}{p(x)}d\mathbf{c}
\end{align}
}

\subsubsection{Loss Derivation.}
Using a variational inference approach, we define an approximate posterior distribution, $q(\mathbf{c} \mid x, c)$, which, like CEM, generates concept embeddings by conditioning on both the input and the concept (as illustrated by the dotted lines in Figure~\ref{fig:pgm}c). This allows for amortized inference, as the true posterior $p(\mathbf{c} \mid x, c)$ is intractable. This approach leads to the derivation of the following Evidence Lower Bound (ELBO) for the log-likelihood of the conditional distribution $p(c, y \mid x)$: 
\begin{align}
\text{log } p(c,y|x)    
\geq& - \underbrace{E_q \left [ \text{log } \frac{q(\mathbf{c}|x,c)}{p(\mathbf{c}|c)}\right ]}_{\text{Pior Matching}} + \underbrace{ \text{log } p(c|x)}_{\text{Concept Loss}} + \underbrace{E_q \left [ \text{log } p(y|\mathbf{c})\right ]}_{\text{Task Loss}} \label{eq:elbo}
\end{align}

A comprehensive derivation of the loss function is provided in Appendix~\ref{app:loss}. The first term in the ELBO is the Kullback-Leibler (KL) divergence between the approximate posterior $q(\mathbf{c} \mid x, c)$ and the prior $p(\mathbf{c} \mid c)$, ensuring their alignment. We refer to this term as \emph{Prior Matching}.  
This alignment is crucial, as it encourages the approximate posterior $q(\mathbf{c} \mid x, c)$—which depends on both the input $x$ and concept predictions $c$—to resemble the prior $p(\mathbf{c} \mid c)$, which is independent of $x$. Maximizing the second and third terms of the ELBO (Concept and Task loss) optimizes both concept and task accuracy. Since the third term involves averaging over concept embeddings sampled from the approximate posterior $q$, we approximate this by using the Monte Carlo method. Specifically, as described in~\cite{kingma2013auto, louizos2017learning}, we employ a large batch size and draw a single sample of $\mathbf{c}$ per data point using the reparameterization trick. All the distributions in the ELBO, besides the prior distribution $p(\mathbf{c}\mid c)$, are parameterized by neural networks.

\subsubsection{Concept Embedding Encoder.}
We assume that each concept $c_j$ is independent of the others. Consequently, we define each concept embedding $\mathbf{c}_j$ as independent of the other concept embeddings and model it as a mixture of two multivariate normal distributions: 
$$
p(\mathbf{c}_j | c_j) = \delta(c_j)\mathcal{N}(\mathbf{c}_j; \mu_j^+, I) + (1-\delta(c_j))\mathcal{N}(\mathbf{c}_j; \mu_j^-, I),
$$
where $\mu_j^+,\mu_j^-\in \mathbb{R}^m$ are learnable embeddings, $I$ is the identity matrix, and $\delta(\cdot)$ represents the Dirac delta function, which evaluates to $1$ if $c_j = 1$ and $0$ otherwise. Here, $\mu_j^+$ corresponds to the expected embedding when the concept is active ($c_j = 1$), while $\mu_j^-$ represents the expected embedding when the concept is inactive ($c_j = 0$). For the sake of simplicity, we define the approximate posterior as a multivariate normal distribution:  
\[
q(\mathbf{c}_j | x, c_j) = \mathcal{N}(\mathbf{c}_j; \hat{\mu}_j(x,c_j), \text{diag}(\sigma_j(x,c_j)))\]
where $\hat{\mu}_j(x,c_j), \text{ }\sigma_j(x,c_j)\in\mathbb{R}^m$.

Given this definition for the prior and the approximate posterior, the \emph{Prior Matching} term can be expressed in a closed-form solution. A detailed derivation of this formulation is presented in Appendix~\ref{app:d_kl}. 
During training, the \emph{Prior Matching} term encourages the approximate posterior $q$ to position the multivariate normal distribution near $\mu_j^+$ when $c_j = 1$ and near $\mu_j^-$ otherwise. This regularization promotes the formation of dense clusters for each concept state, ensuring that each state is represented by a distinct concept embedding: $\mu_j^+$ for $c_j = 1$ and $\mu_j^-$ for $c_j = 0$.
By exploiting this property of V-CEM, we can perform concept embedding intervention, thereby decoupling the concept embedding from the raw input data. 

\subsection{V-CEM Training}  
\label{sec:v-cem_train}
The model is trained to optimize the ELBO by minimizing its negative counterpart. Assuming each concept $c_j$ follows a Bernoulli distribution, the second term in the ELBO reduces to a sum of binary cross-entropy losses, denoted as $L_c$. Similarly, if the task variable $y$ follows a categorical distribution, the third term in ELBO corresponds to the expected cross-entropy loss over $y$, referred to as $L_t$.  

Following standard practices in concept bottleneck models~\cite{zarlenga2022concept}, we introduce a weighting parameter $\lambda_t \in [0,1]$ to balance the task loss $L_t$, allowing for trade-offs between concept learning and task performance. Additionally, a scaling factor $\lambda_p \in [0, \infty)$ is applied to the \emph{Prior Matching} term $L_p$, influencing the model’s regularization. Increasing $\lambda_p$ progressively aligns V-CEM with a CBM, while setting $\lambda_p = 0$ removes constraints on concept embeddings, making the model function like CEM.  

V-CEM is trained by minimizing the following objective function:  
\begin{equation}
L = \frac{1}{k}L_c + \lambda_t L_t + \lambda_p L_p
\end{equation}
where $L_c$ is normalized by the number of concepts $k$. In this work, we set $\lambda_t=0.1$ and $\lambda_p=0.05$. An ablation study exploring the effect of varying $\lambda_p$ on V-CEM’s performance is provided in Appendix~\ref{app:ablation}.  

To enhance the responsiveness of V-CEM to ID interventions, the \emph{RandInt} regularization strategy~\cite{zarlenga2022concept, kim2023probabilistic} is employed during the training phase, performing  random concept embedding interventions with a predefined probability. Additional details about the specific settings of the proposed methodology and the baseline methods are provided in Appendix~\ref{app:training_details}.

\section{Evaluating Concept Representations}
\label{sec:eval}
In order to properly evaluate the model's intervenability in OOD settings, particularly when dealing with concept embeddings, concept accuracy might not be sufficient. In this section, we describe two further metrics that we use for this scope: OOD intervenability and \textit{Concept Representation Cohesiveness (CRC)}. 

\subsubsection{OOD Intervenability.}  
Concept interventions are generally used to assess the intervenability of a model \cite{koh2020concept}, i.e., whether a model's predictions change when concept predictions are modified while keeping other factors constant. 
Model intervenability is normally evaluated ID by replacing concept predictions with concept labels. However, ID concept predictions are often already correct, thus the possibility to obtain a counterfactual prediction is low. Furthermore, for  models relying on concept embeddings, this phenomenon is even more evident as part of the task prediction depends on $x$ rather than $c$. 
Thus, in this paper we evaluate model intervenability OOD.
More specifically, we propose to analyze responsiveness to interventions under varying conditions by progressively adding random noise \(\epsilon \sim N(0, I)\) to the input \(x\). The perturbed input is thus defined as:  
\[
\tilde{x} = (1 - \theta) \cdot x + \theta \cdot \epsilon, \quad \theta \in [0,1]
\]
where \(\theta\) controls the noise intensity. Interventions are applied randomly on misclassified concepts, with an increasing probability \(p_{int} \in [0,1]\).

\subsubsection{Concept Representation Cohesiveness. }  
Concept embeddings allow concept-based models to avoid the performance trade-off due to the CBM concept-bottleneck layer, as they enrich concept representation with sample-based information. Still, it is fundamental that this information represents the concept and not other input features; otherwise we may incur in the so-called “concept leakage” issue \cite{mahinpei2021promises, marconato2022glancenets}, where the concepts encode spurious information related to other concepts. In other words, we would like each point in the concept embedding space \(\mathbf{C}\) to represent a different instantiation of an active or inactive concept. As training a decoder for each concept is non-trivial, in this paper we propose to assess this characteristic through an evaluation of the cohesiveness of the clusters associated with active and inactive concepts.   
More precisely, we compute \textit{CRC}  by splitting all concept embeddings into two clusters according to their concept predictions, and we compute the corresponding silhouette score as follows:
\begin{equation}
\label{eq:crc}    
CRC = \frac{1}{|C|} \sum_{i=0}^{ |C|} s_i(\mathbf{c}_i, c_i)
\end{equation}
where $|C|$ represents the number of concepts and $s_i(\mathbf{c}_i, c_i)$ represents the silhouette coefficient computed for the $i$th concept over concept embedding representation $\mathbf{c}_i$ and considering as clustering labels the concept prediction $c_i$. For further detail on the computation of $s_i$ we refer the reader to Appendix \ref{app:crc}. 
A higher silhouette score indicates a denser and tighter concept embedding space. This, in turn, indicates a model more responsive to OOD concept embedding intervention, as it samples from a denser representation. 

\section{Experimental Evaluation}
\label{sec:exp}

To evaluate V-CEM, we seek to address several key research questions that guide our investigation. Specifically, we aim to answer the following:

\begin{enumerate}[label=(\arabic*)]
    \item Does V-CEM exhibit comparable task performance to Black-box and CEM in ID settings?
    \item Is V-CEM more responsive than concept embedding-based approaches (CEM and Prob-CBM) in OOD scenarios?
    \item How does V-CEM concept representation compare to CBM representation, despite its reliance on concept embeddings?
\end{enumerate}
\subsection{Experimental Setting}

In this section, we outline the experimental setup used to evaluate the performance of V-CEM. Specifically, we present the datasets, the baseline models  and the training details.

\subsubsection{Datasets.}  
We conduct experiments on a diverse set of vision and NLP datasets. For vision, we use MNIST Even/Odd and MNIST Addition, which are derived from the MNIST dataset~\cite{lecun} and involve binary classification and digit-sum prediction tasks, respectively. For these two datasets digits are used as concepts. We conduct experiments also on CelebA~\cite{liu2015faceattributes}, a large-scale facial attribute dataset, where selected attributes serve as concepts and others as prediction targets. For NLP, we experiment with CEBaB~\cite{cebab}, a dataset designed to study causal effects of concepts in sentiment analysis, and IMDB~\cite{imdb}, where movie reviews are classified as positive or negative using interpretable aspects. More details on dataset preprocessing and structure are provided in Appendix~\ref{app:datasets}.

\subsubsection{Baselines.}  
To assess the effectiveness of the proposed methodology, we compare it against several baseline models. For vision tasks, we extract embeddings using a frozen ResNet-34~\cite{resnet}, while for NLP tasks, we use \textit{all-distilroberta-v1}\footnote{We use the pretrained model available at \url{https://huggingface.co/sentence-transformers}.}~\cite{distillbert}. Both backbones are used without fine-tuning to extract embeddings from the input data. All baselines operate on these precomputed embeddings. The compared models include:  
(1) a standard Black-box model, implemented using two consecutive linear layers,  
(2) two variations of CBMs~\cite{koh2020concept}: the first employing a single linear layer to map concepts to the task (CBM+Linear), and the second utilizing two consecutive linear layers (CBM+MLP),  (3) Prob-CBM~\cite{kim2023probabilistic},
(4) CEM~\citep{zarlenga2022concept}. Training details for all models are reported in Appendix~\ref{app:datasets}.

\begin{table}[t]
\caption{The average task accuracy and corresponding standard deviation in ID settings obtained by the various methodologies across different datasets. V-CEM performance are the highest on average when considering concept-based models, surpassing also Black-box performance on three datasets.}
\centering
\begin{tabular}{l|c|c|c|c|c}
\toprule
 & MNIST E/O & MNIST+ & CelebA & CEBaB & IMDB \\
\midrule
Black-box & $98.56$ {\tiny{$\pm 0.01$}} & $67.59$ {\tiny{$\pm 0.57$}} & $\textbf{64.66}$ {\tiny{$\pm 0.07$}} & $\textbf{80.20}$ {\tiny{$\pm 0.25$}} & $86.98$ {\tiny{$\pm 0.48$}} \\
\hline
CBM+Linear & $98.82$ {\tiny{$\pm 0.04$}} & $44.19$ {\tiny{$\pm 1.86$}} & $49.75$ {\tiny{$\pm 0.18$}} & $63.66$ {\tiny{$\pm 3.48$}} & $87.30$ {\tiny{$\pm 0.58$}} \\
CBM+MLP & $98.82$ {\tiny{$\pm 0.13$}} & $68.63$ {\tiny{$\pm 0.72$}} & $51.01$ {\tiny{$\pm 0.51$}} & $72.51$ {\tiny{$\pm 5.88$}} & $86.48$ {\tiny{$\pm 1.88$}} \\
CEM & $98.75$ {\tiny{$\pm 0.07$}} & $69.84$ {\tiny{$\pm 0.91$}} & $64.49$ {\tiny{$\pm 0.08$}} & $80.12$ {\tiny{$\pm 0.14$}} & $86.79$ {\tiny{$\pm 0.77$}} \\
Prob-CBM & $97.38$ {\tiny{$\pm 0.61$}} & $27.31$ {\tiny{$\pm 3.92$}} & $51.64$ {\tiny{$\pm 7.12$}} & $77.86$ {\tiny{$\pm 0.95$}} & $85.90$ {\tiny{$\pm 0.38$}} \\
\hline
V-CEM & $\textbf{98.91}$ {\tiny{$\pm 0.05$}} & $\textbf{73.12}$ {\tiny{$\pm 0.35$}} & $64.49$ {\tiny{$\pm 0.15$}} & $79.62$ {\tiny{$\pm 1.29$}} & $\textbf{87.94}$ {\tiny{$\pm 0.86$}} \\
\bottomrule
\end{tabular}
\label{tab:performance}
\end{table}

\subsection{Results}
\label{sec:res}

The results highlight three key findings: (1) V-CEM outperforms CBMs and Prob-CBM while remaining comparable to CEM and Black-box models in ID settings, (2) it exhibits high responsiveness to interventions in OOD scenarios compared to CEMs and Prob-CBM, and (3) its concept embedding space $\mathbf{C}$ is more cohesive than that of concept embedding-based models.

\begin{figure*}[t]
\centering
\includegraphics[width=\textwidth]{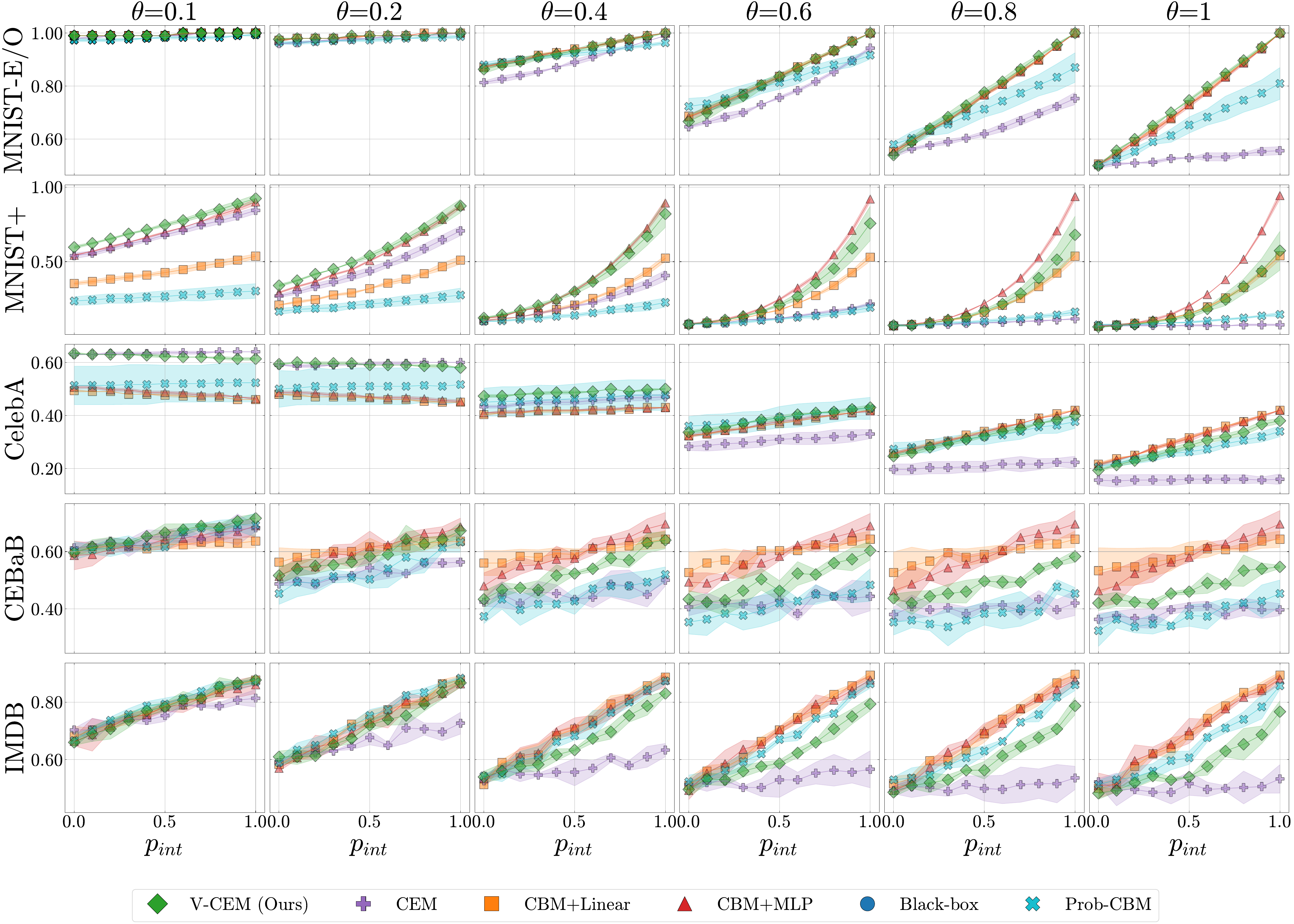}
\caption{
The solid lines represent the mean task accuracy under random interventions at probability $p_{int}$, while the shaded areas indicate the standard deviation of each method. Results are reported across different models and datasets, under varying levels of input noise $\theta \in [0,1]$. The Black-box model is not shown since it does not allow human interventions.
}
\label{fig:intervention}
\end{figure*}

\subsubsection{In-Distribution Performance. }
In Table~\ref{tab:performance}, we present the task accuracy results for the various models evaluated across different datasets in ID settings. The results clearly demonstrate that \textbf{V-CEM consistently outperforms traditional CBMs and Prob-CBM} in average ID performance. This trend is consistent across all datasets and this is particularly evident in MNIST Addition, where V-CEM achieves over 40\% higher task accuracy compared to Prob-CBM and outperforms CBM+Linear by nearly 30\%. Overall, V-CEM achieves ID performance comparable to CEM and the Black-box model while also attaining the highest average accuracy for MNIST E/O, MNIST+, and IMDB. This is achieved while maintaining similar concept accuracy across all models, as reported in Appendix~\ref{app:concept}.

\subsubsection{Intervention Responsiveness.}  
Figure~\ref{fig:intervention} illustrates the task accuracy of various models when human intervention is used to correct misclassified concept predictions under varying levels of input noise $\theta \in [0,1]$, revealing several key insights. 
As anticipated, CEM shows minimal responsiveness to interventions, underscoring a key limitation: its strong dependence on input, which makes it less effective in the presence of distributional shifts. In contrast, \textbf{V-CEM consistently shows greater responsiveness to interventions in OOD settings}, outperforming Prob-CBM, which only surpasses V-CEM in responsiveness for the IMDB dataset. This suggests that V-CEM retains intervention efficacy by more effectively utilizing concept embeddings. Overall, V-CEM demonstrates responsiveness similar to CBMs while achieving superior performance in the ID scenario.

\begin{table}[t]
\caption{The average \textit{CRC} values and their respective standard deviations in ID settings evaluated for all methodologies and datasets. The higher the better. V-CEM values are close to CBMs and always higher than both CEM and Prob-CBM. }
\centering
\begin{tabular}{l|c|c|c|c|c}
\toprule
 & MNIST E/O & MNIST+ & CelebA & CEBaB & IMDB \\
\midrule
CBM+Linear  & $0.99$ {\tiny{$\pm \leq 0.01$}}  & $0.92$ {\tiny{$\pm 0.01$}}       & $0.73$ {\tiny{$\pm 0.01$}}       & $0.70$ {\tiny{$\pm 0.01$}}       & $0.73$ {\tiny{$\pm 0.01$}} \\
CBM+MLP     & $0.99$ {\tiny{$\pm \leq 0.01$}}  & $0.91$ {\tiny{$\pm 0.01$}}       & $0.72$ {\tiny{$\pm 0.01$}}       & $0.71$ {\tiny{$\pm 0.01$}}       & $0.74$ {\tiny{$\pm 0.01$}} \\
CEM         & $0.65$ {\tiny{$\pm 0.01$}}  & $0.65$ {\tiny{$\pm 0.02$}}       & $0.32$ {\tiny{$\pm 0.02$}}       & $0.33$ {\tiny{$\pm 0.03$}}       & $0.45$ {\tiny{$\pm 0.04$}} \\
Prob-CBM    & $0.73$ {\tiny{$\pm 0.01$}}  & $0.59$ {\tiny{$\pm 0.02$}}       & $0.31$ {\tiny{$\pm 0.03$}}       & $0.41$ {\tiny{$\pm 0.05$}}       & $0.50$ {\tiny{$\pm 0.02$}} \\
\midrule
V-CEM       & $0.98$ {\tiny{$\pm 0.01$}}  & $0.85$ {\tiny{$\pm 0.02$}}       & $0.41$ {\tiny{$\pm 0.03$}}       & $0.59$ {\tiny{$\pm 0.02$}}       & $0.67$ {\tiny{$\pm 0.02$}} \\
\bottomrule
\end{tabular}
\label{tab:silhouette}
\end{table}

\begin{figure*}[h]
\centering
\includegraphics[width=\textwidth]{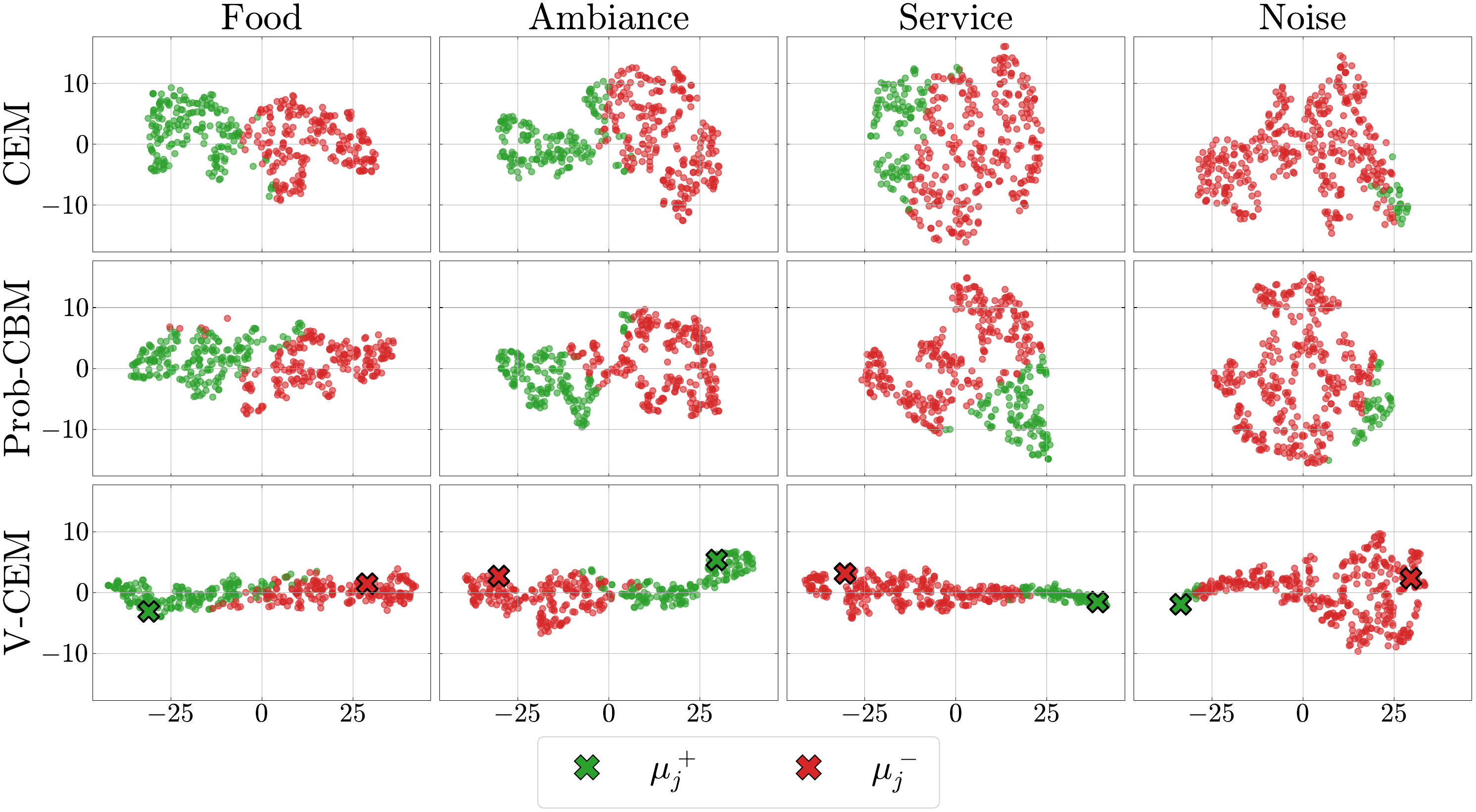}
\caption{2D t-SNE visualization of the concept embedding space $\mathbf{c}$ for the CEBaB dataset, comparing V-CEM, Prob-CBM and CEM. V-CEM concept representation is much denser than the ones of CEM and Prob-CEM.}
\label{fig:cebab_space}
\end{figure*}

\subsubsection{Concept embedding space evaluation.}  
As outlined in Section~\ref{sec:exp}, to further investigate why V-CEM outperforms CEMs in terms of intervention responsiveness while maintaining performance comparable to CBMs in OOD settings, we propose to analyze the cohesiveness of the concept embedding space.

Ideally, each point in the concept embedding space should correspond to a distinct instance of an active or inactive concept. 
Specifically, for each concept, we here identify two clusters and compute the \textit{CRC} score across all concepts for each model and dataset. Table~\ref{tab:silhouette} presents the results, showing that \textbf{V-CEM’s concept embedding space is more cohesive than that of concept embedding based models} (CEM and PRob-CBM) while remaining comparable to CBMs. Additionally, Figure~\ref{fig:cebab_space} provides a 2D t-SNE visualization comparing the concept embedding spaces of CEM, Prob-CBM and V-CEM for different concepts in the CEBaB dataset, further illustrating this effect.

\section{Related Works}
\label{sec:related}
C-XAI~\cite{poeta2023concept} has gained prominence as a solution to the growing demand for machine learning models that provide explanations in human-understandable terms~\cite{rudin2019stop}. Unlike traditional feature-based approaches, C-XAI emphasizes associating a model’s behavior with human-interpretable concepts, offering a more intuitive and accessible way to understand model decisions.

A foundational approach in this domain is Testing with Concept Activation Vectors (T-CAVs), introduced in~\cite{kim2018interpretability}, which leverages directions in the latent space to measure, post-hoc, a model’s sensitivity to predefined human-understandable concepts.
In contrast, explainable-by-design models incorporate interpretability directly into their architecture. A prominent example is CBM~\cite{koh2020concept, ciravegna2023logic, alvarez2018towards}, which integrates supervised concepts as intermediate representations within the prediction pipeline, enabling more transparent and controllable decision-making.
This integration facilitates direct intervention and correction of errors, thus enabling a more interactive approach to model understanding.

Additionally, Concept Embedding Models (CEMs)~\cite{zarlenga2022concept} push the boundaries of the interpretability-performance trade-off by incorporating concept embeddings into the learning process, enabling richer representations while maintaining concept-based explanations. Building on this approach, Prob-CBM\cite{kim2023probabilistic} employs concept embeddings to capture uncertainty in concept predictions, offering explanations that incorporate both the concept and its associated uncertainty.

However, a critical challenge remains: ensuring robustness in OOD scenarios. OOD generalization is essential in machine learning, as it determines a model’s ability to maintain reliable performance when encountering data that deviates from the training distribution.

Methodologies such as Outlier Exposure~\cite{hendrycks2018deep} and ODIN~\cite{liang2017enhancing} aim to enhance the detection and management of OOD samples during inference. Additionally, studies like~\cite{gulrajani2020search} underscore the necessity of benchmarking OOD performance through meticulously designed experimental protocols. More recently, other works~\cite{vishwakarma2023human, bai2024aha} have demonstrated that the decline in the performance of deep learning models following deployment in real-world applications can be mitigated by incorporating human assistance to support OOD generalization. Motivated by this approach, we seek to investigate the potential of leveraging the interveneability characteristic of concept-based models to enable human interventions under OOD conditions. The interplay between OOD generalization and concept-based explainability remains a relatively unexplored yet promising research direction. Integrating these domains could pave the way for more resilient and interpretable models that provide actionable explanations, even in challenging and unforeseen scenarios.

\section{Conclusion}
\label{sec:conc}

We introduced V-CEM, a model that achieves ID performance comparable to Black-box and CEM while maintaining strong responsiveness to interventions in both ID and OOD settings. We have shown that its improved performance compared to other concept embedding based models originates from the cohesiveness of its concept embedding space which is ensured by the generative process that is conditioned  on the concept prediction only.

Although V-CEM demonstrates strong responsiveness in OOD scenarios, it lacks an inherent mechanism for identifying OOD samples. Implementing such a mechanism would be beneficial, as it could assist human intervention by highlighting concepts associated with samples that deviate from those the model encountered during training. This could improve the model's ability to flag and address potential OOD instances, enhancing its overall reliability and reducing the risk of misclassification. Moreover, V-CEM was evaluated exclusively on OOD scenarios generated by introducing random noise into the input. Additional experiments are required to assess its performance on other types of distributional shifts.

\subsubsection{Future works.}
Future research directions include extending V-CEM to handle multimodal inputs, allowing the model to integrate and process information from multiple data sources effectively, thereby creating shared and aligned concept embeddings across modalities. 
Another promising avenue is the incorporation of generative models as decoders for concepts, leveraging their capabilities to create concept visualizations from V-CEM cohesive concept embedding space.
Finally, V-CEM models concepts independently from each other. In scenarios where the presence of a concept significantly affects other concepts, it may be opportune to explicitly model this dependency. Merging V-CEM with the strategy suggested in \cite{dominici2024causal, de2025causally} might offer a method for accomplishing this.

\section*{Acknowledgement}
The research leading to these results has been funded by the Italian Ministry of University as part of the 2022 PRIN Project ACRE (AI-Based Causality and Reasoning for Deceptive Assets - 2022EP2L7H).

%
%
%

\newpage
\appendix 
\renewcommand{\thesection}{\arabic{section}} 
\section*{Appendix}
\section{Loss derivation}
\label{app:loss}
This appendix provides the complete derivation of the loss function that V-CEM is trained to approximate:

{\small
\begin{align}
    \text{log } p(c,y|x) =& \text{ log }\int_{\mathbf{C}} \frac{p(x,c,\mathbf{c},y)}{p(x)}d\mathbf{c} \label{eq:5}\\ 
    =& \text{ log }\int_\mathbf{C}p(c|x)p(\mathbf{c}|c)p(y|\mathbf{c}) d\mathbf{c} \label{eq:6}\\ 
    =& \text{ log } \int_\mathbf{C} \frac{q(\mathbf{c}|x,c)}{q(\mathbf{c}|x,c)}p(c|x)p(\mathbf{c}|c)p(y|\mathbf{c}) d\mathbf{c}\label{eq:7}\\ 
    =& \text{ log } E_q \left [ \frac{p(c|x)p(\mathbf{c}|c)p(y|\mathbf{c})}{q(\mathbf{c}|x,c)}\right] \label{eq:8}\\ 
    \geq& E_q \left [ \text{log }\frac{p(c|x)p(\mathbf{c}|c)p(y|\mathbf{c})}{q(\mathbf{c}|x,c)}\right] \label{eq:9}\\ 
    =& - E_q \left [ \text{log } \frac{q(\mathbf{c}|x,c)}{p(\mathbf{c}|c)}\right ] + \text{log } p(c|x) + E_q \left [ \text{log } p(y|\mathbf{c})\right ] \label{eq:10}
\end{align}
}
We begin by re-expressing the target conditional probability $p(x,y \mid \mathbf{c})$ through marginalization over $\mathbf{C}$ and factorizing the joint distribution $p(x,c,\mathbf{c},y)$ according to the generative process illustrated in Figure~\ref{fig:pgm}c. Next, to amortize inference we introduce an approximate posterior distribution $q(\mathbf{c}|x, c)$ (Eq.~\ref{eq:7}). By applying Jensen's inequality, we obtain a lower bound on the log-likelihood, known as the ELBO, as shown in Eq.~\ref{eq:9}. Finally, Eq.~\ref{eq:10} expands the ELBO into three terms: the first term is the negative KL divergence between $q(\mathbf{c}|x,c)$ and $p(\mathbf{c}|c)$, which measures the difference between the approximate posterior and the true prior; the second term is the log-likelihood of $c$, and the third term is the expected log-likelihood of $y$.

\section{Prior matching formulation}
\label{app:d_kl}
An important assumption we make, which is a standard assumption for concept based methodologies, is that the different concepts, and therefore the different concepts embeddings, are independent one another. Therefore, $q(\mathbf{c}|x,c)=\prod_{j=1}^k q(\mathbf{c}_j|x,c_j)$ and $p(\mathbf{c}|c)=\prod_{j=1}^k p(\mathbf{c}_j|c_j)$. This allows to rewrite the \emph{Prior Matching} term as the sum of KL divergences between the approximate posterior and the true prior of each concept:

{\small
\begin{align}
E_q \left [ \text{log } \frac{q(\mathbf{c}|x,c)}{p(\mathbf{c}|c)}\right ] =& \int_\mathbf{C}  q(\mathbf{c}|x,c) \text{log } \frac{q(\mathbf{c}|x,c)}{p(\mathbf{c}|c)} d\mathbf{c} \label{eq:11}\\
=& \sum_{j=1}^k \int_\mathbf{C} \prod_{i=1}^{k}q(\mathbf{c}_i|x,c_i) \text{ log } \frac{q(\mathbf{c}_j\mid x, c_j)}{p(\mathbf{c}_j\mid c_j)} d\mathbf{c} \label{eq:12}\\ 
=& \sum_{j=1}^k \int_\mathbf{C} q(\mathbf{c}_j|x,c_j) \text{ log } \frac{q(\mathbf{c}_j\mid x, c_j)}{p(\mathbf{c}_j\mid c_j)} d\mathbf{c} \label{eq:13}\\ 
=& \sum_{j=1}^k E_q \left [ \text{log } \frac{q(\mathbf{c}_j|x,c_j)}{p(\mathbf{c}_j|c_j)}\right ] \label{eq:14}
\end{align}
}

The prior is modeled as a mixture, governed by the function $ \delta(\cdot) $, which selects the appropriate normal distribution based on the value of $ c_j $. As a result, the KL divergence is computed differently depending on whether $ c_j $ is active or inactive. When $ c_j = 1 $, it quantifies the divergence between the approximate posterior and the corresponding normal distribution in the prior for $ c_j = 1 $. Similarly, when $ c_j = 0 $, it measures the divergence between the approximate posterior and the prior distribution associated with $ c_j = 0 $. Defining

\[
\mu_j =
\begin{cases} 
\mu_j^+ & \text{if } c_j = 1, \\
\mu_j^- & \text{if } c_j = 0
\end{cases}
\]

allows to rewrite the \emph{Prior Matching} term as:

{\small
\begin{align}
    E_q \left [ \text{log } \frac{q(\mathbf{c}|x,c)}{p(\mathbf{c}|c)}\right ] &= \frac{1}{2}\sum_{j=1}^k\left [||\hat{\mu}_j(x,c) - \mu_j||^2 + \sum_{z=1}^m \sigma_{jz}^2(x,c) - m - \sum_{z=1}^m \text{log }\sigma_{jz}^2(x,c)
    \right ] \nonumber
\end{align}
}

where $\sigma^2_{jz}(x,c)$  denotes the variance of the concept embedding $j$ for the latent dimension $z$.

\section{Concept Representation Cohesiveness}
\label{app:crc}
In our manuscript we introduce a novel metric to compute the Concept Representation Cohesiveness, a metric to comprehend how spread the representation are in the concept space which is particularly useful to assess how prone a model is to concept leakage and in turn how likely we can correctly perform concept intervention also OOD. Recalling from Section  \ref{sec:eval}, Equation \ref{eq:crc}, we defined CRC as:
\[
CRC = \frac{1}{|C|} \sum_{i=0}^{ |C|} s_i(\mathbf{c}_i, c_i)
\]
More specifically, we now define how to compute $s_i$ (here and in the following we drop the dependency from $\mathbf{c}_i$, $c_i$): 
\[
\qquad s_i= \frac{1}{2}\left(\frac{b_i^+ - a_i^+}{\text{max}(b_i^+, a_i^+)} + \frac{b_i^- - a_i^-}{\text{max}(b_i^-, a_i^-)}\right),  
\]
\[
a_i^+ = \frac{1}{|\mathcal{C}_i^+|}\sum_{j\in \mathcal{C}_i^+} \frac{1}{|\mathcal{C}_i^+-1|}\sum_{k\in\mathcal{C}_i^+, k\neq j}{||\mathbf{c}_{ij} - \mathbf{c}_{ik}||_1}
\]
\[
b_i^+ = \frac{1}{|\mathcal{C}^+ - 1|} \sum_{j\in \mathcal{C}_i^+} \frac{1}{|\mathcal{C}^-|} \sum_{k\in \mathcal{C}_i^-} ||\mathbf{c}_{ij} - \mathbf{c}_{ik}||_1 
\]
and where $\mathcal{C}_i^+ \mathcal{C}_i^-$ are the set of sample indexes associated to positive and negative concept prediction for concept $i$ and are thus computed as: \(\mathcal{C}_i^+ = \mathbbm{1}_{c_i> 0.5}\) and \(\mathcal{C}_i^- = \mathbbm{1}_{c_i \leq 0.5}\).

\section{Ablation on $\lambda_p$ variation}
\label{app:ablation}

In our manuscript we introduce a scaling factor $\lambda_p \in [0, \infty)$ to regulate the \emph{Prior Matching} term, allowing fine-grained control over the model’s behavior. Increasing $\lambda_p$ progressively aligns V-CEM with a standard CBM, whereas setting $\lambda_p = 0$ eliminates constraints on concept embedding generation, making the model function similarly to a CEM. In this appendix we show how modifying $\lambda_p$ modifies the model performance.

\begin{figure*}[h]
\centering
\includegraphics[width=0.5\textwidth]{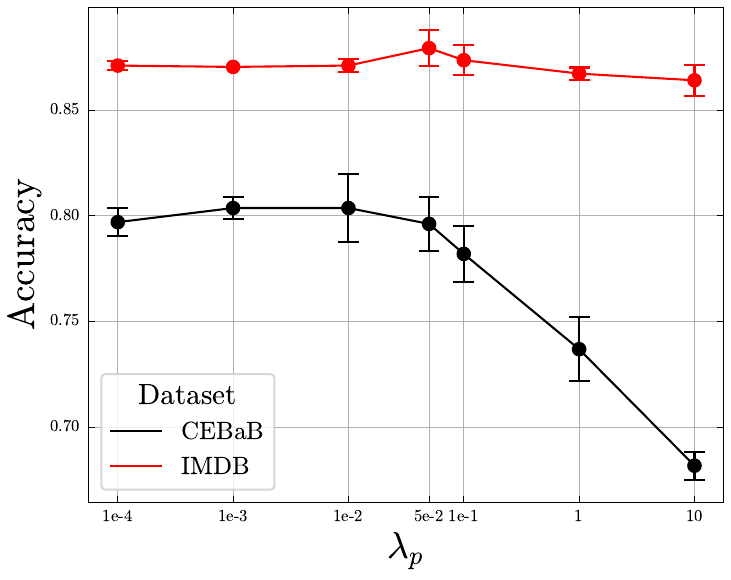}
\caption{Variation in V-CEM's ID accuracy across different values of $\lambda_p$ on the CEBaB and CelebA datasets.}
\label{fig:ablation_acc}
\end{figure*}

In Figure~\ref{fig:ablation_acc}, we report the ID performance of V-CEM on CEBaB and IMDB as an example of performance datasets when modifying $\lambda_p$. The observed transition aligns with expectations: for $\lambda_p = 0$, the model achieves good performance similar to CEM, while increasing $\lambda_p$ leads to performance degradation, making it more similar to that of CBMs. 

\begin{figure*}[h]
\centering
\includegraphics[width=\textwidth]{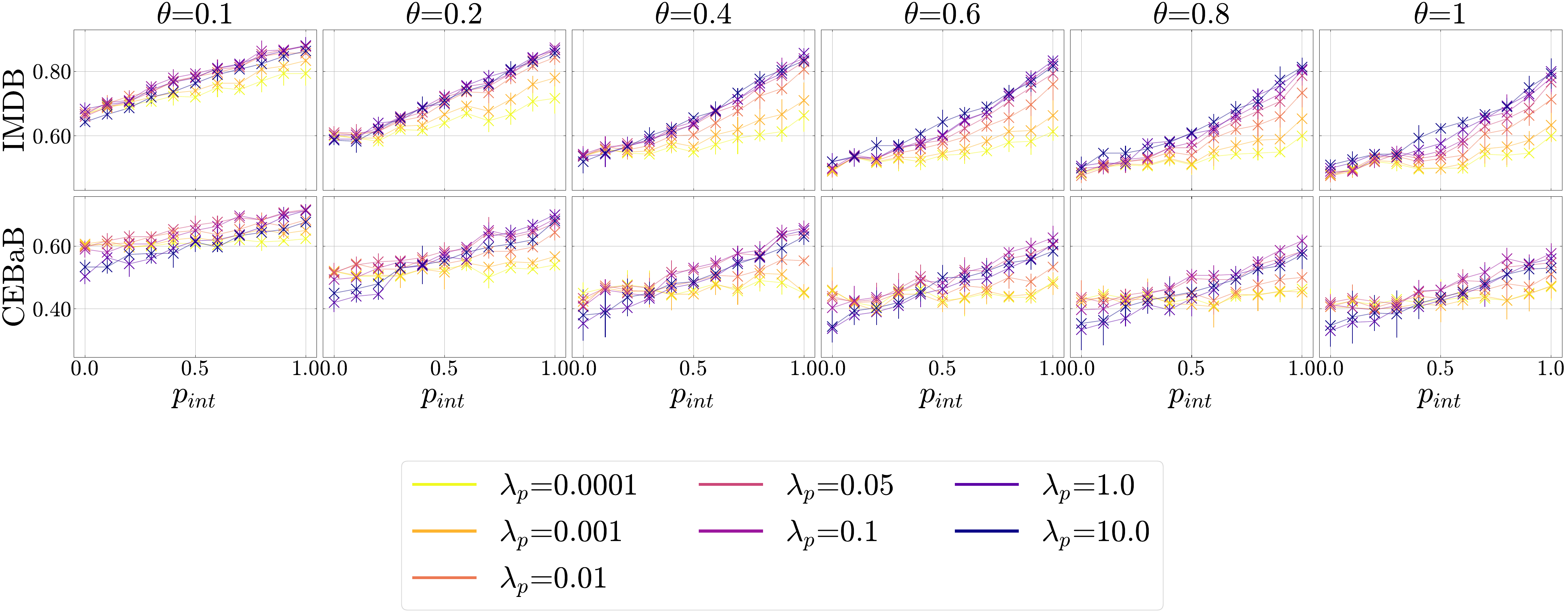}
\caption{Impact of interventions on V-CEM's OOD accuracy across different $\lambda_p$ values.}
\label{fig:ablation_int}
\end{figure*}

Similar results can be observed in Figure~\ref{fig:ablation_int}, where for low values of $\lambda_p$, responsiveness to interventions is weaker—a characteristic typical of CEM—while it improves as $\lambda_p$ increases, approaching the responsiveness of CBMs. To balance both in-distribution performance and responsiveness to interventions, we set $\lambda_p = 0.05$ in this manuscript.  

\section{Dataset details}
\label{app:datasets}
In this appendix, we provide additional details on the datasets used to evaluate the performance of the tested models, followed by a description of the training procedure.

\subsection{Datasets}
\subsubsection{MNIST Even/Odd.}  
MNIST Even/Odd is a binary classification dataset derived from MNIST, where digits, used as concept labels, are categorized as either even or odd. It consists of \num{60000} training images and \num{10000} test images, each of size $28\times28$ and in grayscale. All images were converted to a three-channel format and rescaled to $224\times 224$. 

\subsubsection{MNIST Addition.}  
MNIST Addition is constructed by pairing two MNIST digits (used as concepts) and assigning a label equal to the sum of their individual values. The dataset retains the original MNIST structure, containing \num{60000} training samples and \num{10000} test samples. Each input is a grayscale image formed by concatenating two MNIST digits side by side. Alslo in this case, images were converted to a three-channel format and rescaled to $224\times 224$.  

\subsubsection{CelebA.}  
CelebA is a large-scale facial attribute dataset containing over \num{200000} images of celebrities, each of size $178\times218$. The dataset is divided into training, validation, and test sets.  
We use the following attributes as concepts: \texttt{No Beard}, \texttt{Young}, \texttt{Attractive}, \texttt{Mouth Slightly Open}, \texttt{Smiling}, \texttt{Wearing Lipstick}, and \texttt{High Cheekbones}, as they are the most balanced attributes in the dataset. The task is a multi-class classification problem, where the goal is to predict the attributes \texttt{Wavy Hair}, \texttt{Black Hair}, and \texttt{Male}.  
All images are already in RGB format and are rescaled to $224\times 224$.  

\subsubsection{CEBaB.}  
CEBaB is a dataset designed to study the causal effects of real-world concepts on NLP models. It includes short restaurant reviews annotated with sentiment ratings at both the overall review level (positive, neutral, and negative reviews) and for four dining experience aspects, which are used as concept labels: \texttt{Good Food}, \texttt{Good Ambiance}, \texttt{Good Service}, and \texttt{Good Noise}.  

\subsubsection{IMDB.}  
\sloppy
The IMDB dataset consists of \num{50000} movie reviews labeled as either positive or negative. To predict the overall review sentiment, we use four interpretable concepts: \texttt{Acting}, \texttt{Cinematography}, \texttt{Emotional arousal}, and \texttt{Storyline}. \\\\
For datasets that do not provide a validation set, we randomly removed 10\% of the training data to create a validation set.

\subsection{Training details}
\label{app:training_details}
All models were trained up to $500$ epochs, employing an early stopping criterion with a patience of $20$ epochs. 
The Adam optimizer was used along with a learning rate scheduler that reduced the learning rate by a factor of \( 0.1 \) every 100 epochs. The initial learning rate was dataset-specific: \( 2e{-3} \) for MNIST Even/Odd and MNIST Addition, \( 1e{-4} \) for CelebA, \( 5e{-4} \) for CEBaB, and \( 1e{-2} \) for IMDB.
All baseline models were trained with default hyperparameters. Both Prob-CBM and CEM were trained following the \emph{RandInt} technique proposed in \cite{zarlenga2022concept}, setting it to $0.25$ for CEM and to $0.5$ for Prob-CBM, as suggested in the respective papers. To ensure a fair comparison across different methodologies, we applied the \emph{RandInt} technique during the training of CBM+MLP, CBM+Linear, and V-CEM. Specifically, we set the intervention probability to $0.25$ for these approaches. For V-CEM, random interventions were introduced starting from the 20th epoch for the CelebA dataset (given the larger size of the training-set), while for all other datasets, they were applied from the 3rd epoch onward.

For the V-CEM model, the \emph{Prior Matching} term was scaled using a factor of $\lambda = 0.05$. As for Prob-CBM and CEM, we used a concept embedding dimension of $16$. Each model was trained using three different random seeds.

\section{Concept accuracy}
\label{app:concept}
In this appendix, we report the concept accuracy values for all models and datasets. The results reported in Table~\ref{tab:concept} confirm that, in terms of concept accuracy, the performance of all models is comparable, with V-CEM being on average the best (despite overlapping standard deviations).

\begin{table}[h]
\caption{Concept accuracy comparison across different datasets in ID settings.}
\centering
\begin{tabular}{l|c|c|c|c|c}
\toprule
 & MNIST E/O & MNIST+ & CelebA & CEBaB & IMDB \\
\midrule
CBM+Linear & $99.44$ {\tiny{$\pm 0.01$}} & $95.24$ {\tiny{$\pm 0.00$}} & $83.02$ {\tiny{$\pm 0.03$}} & $80.33$ {\tiny{$\pm 0.74$}} & $84.41$ {\tiny{$\pm 0.05$}} \\
CBM+MLP & $99.46$ {\tiny{$\pm 0.01$}} & $95.08$ {\tiny{$\pm 0.04$}} & $82.91$ {\tiny{$\pm 0.05$}} & $80.81$ {\tiny{$\pm 0.79$}} & $84.49$ {\tiny{$\pm 0.12$}} \\
CEM & $99.35$ {\tiny{$\pm 0.03$}} & $94.91$ {\tiny{$\pm 0.05$}} & $82.77$ {\tiny{$\pm 0.04$}} & $79.51$ {\tiny{$\pm 0.17$}} & $83.30$ {\tiny{$\pm 0.18$}} \\
Prob-CBM & $99.18$ {\tiny{$\pm 0.07$}} & $95.08$ {\tiny{$\pm 0.09$}} & $82.93$ {\tiny{$\pm 0.03$}} & $80.70$ {\tiny{$\pm 0.66$}} & $83.48$ {\tiny{$\pm 0.41$}} \\
\hline
V-CEM & $99.49$ {\tiny{$\pm 0.00$}} & $95.22$ {\tiny{$\pm 0.09$}} & $83.04$ {\tiny{$\pm 0.01$}} & $80.37$ {\tiny{$\pm 0.52$}} & $84.16$ {\tiny{$\pm 0.19$}} \\
\bottomrule
\end{tabular}
\label{tab:concept}
\end{table}

\section{ID Interventions}

\begin{figure*}[h]
\centering
\includegraphics[width=\textwidth]{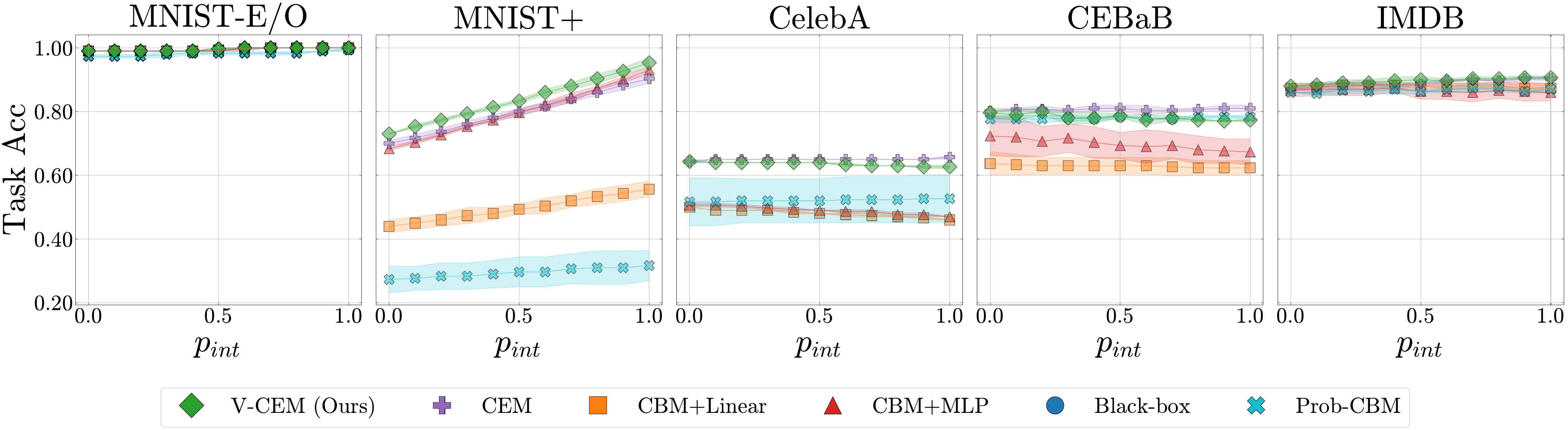}
\caption{Mean and standard deviation of task accuracy with random interventions at probability $p_{int}$ across different models and datasets without noise (ID settings).}
\label{fig:int_id}
\end{figure*}

In addition to demonstrating strong responsiveness to interventions in OOD settings, V-CEM maintains high accuracy even when interventions occur in ID settings. As shown in Figure~\ref{fig:int_id}, the performance of the various models remains stable across different datasets. This stability is primarily attributed to the high concept accuracy (Table~\ref{tab:concept}) achieved by these models, which limits the potential for further improvement following interventions. Notably, in MNIST+, accuracy increases linearly with the intervention probability ($p_{int}$) for all the methodologies. Conversely, for CBM+MLP and CBM+Linear on the CEBaB and CelebA datasets, performance slightly declines post-intervention, likely due to the lower concept accuracy in these datasets (approximately $80\%$). This observation highlights the greater robustness of concept embedding methodologies to interventions in such scenarios.

\bibliographystyle{splncs04}
\bibliography{main}

\end{document}